%% file: main.tex
\documentclass{article}

\usepackage{microtype}
\usepackage{graphicx}
\usepackage{subfigure}
\usepackage{booktabs} 
\usepackage{lipsum}
\usepackage{gensymb}
\usepackage{bbm}
\usepackage{xcolor}
\usepackage{amsmath} 
\usepackage{url}
\usepackage{breakurl}

\usepackage[breaklinks]{hyperref}

\usepackage{hyperref}
\usepackage{balance}

\usepackage[accepted]{icml2021}

\icmltitlerunning{HER Accelerates PPO}

\begin{document}

\input{source/__front_matter}

\begin{abstract}
\input{source/_c_abstract}
\end{abstract}

\input{source/_d_introduction}
\input{source/_e_methods}

\input{source/_f_results}
\input{source/_g_discussion}
\input{source/_h_conclusions}
\balance

\section*{Acknowledgements}

\input{source/_a_acknowledgements}

\cleardoublepage
\bibliography{bib}
\bibliographystyle{icml2021}

\appendix
\renewcommand\thefigure{\thesection\arabic{figure}}
\setcounter{figure}{0}

\cleardoublepage
\input{source/_i_supplement}

\end{document}

%% file: source/__front_matter.tex
\twocolumn[
\icmltitle{Hindsight Experience Replay Accelerates Proximal Policy Optimization}

\icmlsetsymbol{equal}{*}

\begin{icmlauthorlist}
\icmlauthor{Douglas C. Crowder}{cec}
\icmlauthor{Darrien M. McKenzie}{afh}
\icmlauthor{Matthew L. Trappett}{asc}
\icmlauthor{Frances S. Chance}{cec}
\end{icmlauthorlist}

\icmlaffiliation{cec}{Cognitive and Emerging Computing, Sandia National Laboratories, Albuquerque, New Mexico, USA}
\icmlaffiliation{afh}{Autonomy for Hypersonics, Sandia National Laboratories, Albuquerque, New Mexico, USA}
\icmlaffiliation{asc}{Autonomous Sensing and Control, Sandia National Laboratories, Albuquerque, New Mexico, USA}

\icmlcorrespondingauthor{Douglas C. Crowder}{dccrowd@sandia.gov}

\icmlkeywords{Reinfocement Learning, Hindsight Experience Replay}

\vskip 0.3in
]



\printAffiliationsAndNotice{} 

%% file: source/_c_abstract.tex
Hindsight experience replay (HER) accelerates off-policy reinforcement learning algorithms for environments that emit sparse rewards by modifying the goal of the episode \textit{post-hoc} to be some state achieved during the episode.  Because \textit{post-hoc} modification of the observed goal violates the assumptions of on-policy algorithms, HER is not typically applied to on-policy algorithms.  Here, we show that HER can dramatically accelerate proximal policy optimization (PPO), an on-policy reinforcement learning algorithm, when tested on a custom predator-prey environment.

%% file: source/_d_introduction.tex
\section{Introduction}

Reinforcement learning (RL) has been remarkably successful at solving complex tasks \cite{mnih2013playing, mnih2015human, lillicrap2015continuous}.  However, RL can be quite sample inefficient, especially when provided with sparse rewards that require long rollouts to achieve.  To make RL more sample efficient, it is common to shape rewards to make them more informative.  However, the process of shaping rewards can be tedious, and improperly shaped rewards will result in convergence to sub-optimal policies \cite{ng1999policy}.

To make RL more sample-efficient while using sparse rewards, it is possible to use hindsight experience replay (HER) \cite{andrychowicz2017hindsight}, which modifies the goal of an episode \textit{post-hoc} to be some state that was achieved during the episode.  HER is particularly helpful in environments where random walks are inefficient, such as high-dimensional environments and systems that have redundant, antagonistic actuators \cite{crowder2021hindsight}.

However, the use of HER has traditionally been limited to off-policy algorithms, such as DQN \cite{mnih2013playing}, DDPG \cite{lillicrap2015continuous}, TD3 \cite{fujimoto2018addressing}, and SAC \cite{haarnoja2018soft} because on-policy algorithms make the assumption that data used to update the policy was generated by the current policy, which generally prohibits \textit{post-hoc} modification of the observed goal \cite{sutton2018reinforcement}.  However, because on-policy algorithms can exhibit greater stability and clock-time efficiency, the application of HER to on-policy algorithms, such as proximal policy optimization (PPO) \cite{schulman2017proximal}, is of great interest \cite{plappert2018multi}.

\section{Previous Work}

Several previous works have made progress in applying HER to on-policy algorithms.  \cite{rauber2017hindsight} derived a version of HER for policy gradient methods and applied their algorithm in environments that had discrete action spaces. \cite{zhang2019hindsight} combined HER with TRPO.  \cite{dai2020episodic} used HER to improve self-imitation learning \cite{oh2018self} by increasing the number of ``expert'' trajectories available for training the agent via PPO.  Notably, \cite{dai2020episodic} applied these HER-enabled methods in environments that had continuous action spaces.  The methods in \cite{dai2020episodic} relied on both self-imitation learning and a novel transition filtering method to select transitions to maximize the rewards earned by the agent.

Here, we show that a naive, vanilla combination of PPO and HER, without trajectory filtering or self-imitation learning, is sufficient to solve environments that have continuous action spaces.  Additionally, we show that the combined PPO-HER algorithm has comparable sample efficiency to offline RL algorithms, while also being more clock-time efficient.  These results were achieved in a series of predator-prey environments.  We also tested the ability of PPO-HER to generalize to other environments such as the  Fetch environments \cite{plappert2018multi}, and we show that progress must be made before PPO-HER can be applied in these environments.

%% file: source/_e_methods.tex
\section{Task}

\begin{figure*}
    \vskip 0.2in
    \centering
    \centerline{\includegraphics[width=\textwidth]{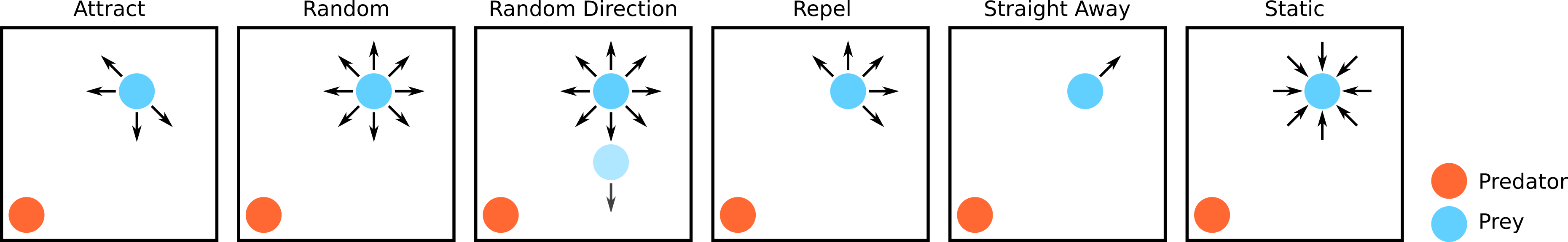}}
    \caption{Custom predator-prey environment.}
    \label{fig:task}
    \vskip -0.2in
\end{figure*}

\subsection{Predator-Prey Environments}
To characterize the behavior of PPO-HER, we created a novel series of predator-prey environments, as shown in Figure \ref{fig:task}.  In these environments, a Predator agent learned via RL to intercept a Prey agent, which executed one of the following predetermined policies:
\begin{enumerate}
    \item \textbf{Attract}: The Prey moved towards the predator at an angle of $\pm [6\degree, 90\degree]$.
    \item \textbf{Random}: The Prey moved in a random direction at each time step.
    \item \textbf{Random Direction}: The Prey moved in the same random direction during an entire episode.
    \item \textbf{Repel}: The Prey moved away from the predator at an angle of $[-90\degree, 90\degree]$.
    \item \textbf{Straight Away}: The Prey moved away from the predator at an angle of exactly $0\degree$.
    \item \textbf{Static}: The Prey remained stationary.
\end{enumerate}

The environments were parameterized by several variables: $D$ the number of physical dimensions, $S$ the length of a single dimension (environments were hypercubes), and $I$ the interception distance.  Unless otherwise specified, $D = 3$, $S = 10$, $I = 1$.  By changing, $D$, $S$, or $I$, we could make the environments more difficult as necessary.

The Predator and Prey spawned into the environment according to 2 policies:
\begin{enumerate}
    \item \textbf{Random}: The Prey and the Predator locations were chosen from a uniform distribution across the entire workspace.
    \item \textbf{Apart}: The Predator always spawned at $[0, 0, \ldots, 0]$ (the zero vector).  The Prey always spawned at $[0, 0, \ldots, S/2]$.
\end{enumerate}

The Predator moved at a maximum speed of $v_{\text{pred}} = |\mathbbm{1}^D|$, the 2-norm of a D-dimensional vector of 1's. The Prey always moved at a speed of $v_{\text{prey}} = v_{\text{pred}} / 2$, accept for the Random policy, when it moved at $v_{\text{prey}} \in [0,  v_{\text{pred}} / 2]$.  The boundaries of the environment were treated as impenetrable walls.

At each time step, the Predator observed the $D$-dimensional position and velocity vectors for itself and the Prey.  The goal of each episode was to reach the Prey position.  The Predator controlled its own $D$-dimensional velocity vector.  If the Predator intercepted the Prey, it received a reward of 1, and the episode ended.  After 20 environment steps, the episode terminated, and the Predator received a reward of -1.  For all non-terminal time steps, the Predator received a reward of 0.  These environments span a large range of difficulties: the Attract environment is trivial to solve, whereas the Straight Away environment is nearly impossible to solve with random exploration, especially as the number of dimensions increases.  We repeated each simulation condition 12 times.

\subsection{Fetch Environments}
To compare with previous studies, we leveraged the Fetch environment \cite{plappert2018multi}, which involves learning to manipulate a robotic arm to move a block.  We replicated each simulation 4 times.  To encourage convergence, we set the L2 regularization parameter to 0.0001.

\section{Reinforcement Learning}
We used high-quality implementations of PPO, SAC, and SAC-HER, provided by \cite{stable-baselines3}, with the default parameters, unless otherwise specified.  For SAC, we decreased the size of the neural networks to 2 layers of 64 nodes/layer to match the neural network topology of PPO.  For PPO, we set target\_kl $= 0.05$ to stabilize learning.

To implement PPO-HER, we took a naive approach of simply resampling goals from the rollout buffer, as done in \cite{andrychowicz2017hindsight} and then recalculating the log probabilities of the actions based on the current policy.  The observation vector at time $t$, $o_t$, was composed of a state $s_t$ and a goal $g_t$, $o_t = \{s_t, g_t\}$.  HER creates an alternative observation by reassigning the goal $g_t$ to be some goal state achieved during the episode, $g'_t$, yielding $o'_t = \{s_t, g'_t\}$.  PPO training requires knowledge about the log probability of choosing action $a_t$ at time $t$: $\text{log}(p(a_t | o_t))$.  When using HER, we re-calculated this probability as $\text{log}(p(a_t | o'_t))$.  We did not rely on self-imitation learning or transition filtering as performed in previous works \cite{dai2020episodic}.

There are several hyperparameters that control HER including the ``strategy'' parameter, which determine how goals are selected, and the ``$k$'' parameter, which determines the ratio of HER transitions to normal transitions that are added to the buffer \cite{andrychowicz2017hindsight}.  The ``strategy'' and ``$k$'' parameters were optimized separately for SAC and PPO using the Static Prey policy and then applied to all other simulation conditions.  In both cases, the optimal HER resampling strategy was found to be ``final,'' which does not require a $k$ parameter.  The ``final'' strategy involves substituting the desired goal with the position that the agent achieved at the final timestep.  Results of the hyperparameter search for PPO are shown in Supplementary Figure \ref{fig:herType}. To calculate clock time, all experiments were provided with the same hardware resources (Intel E5-2683v3 2.00GHz CPUs).

%% file: source/_f_results.tex
\section{Results}

\subsection{Predator-Prey}

\begin{figure*}
    \centering
    \includegraphics[width=\textwidth]{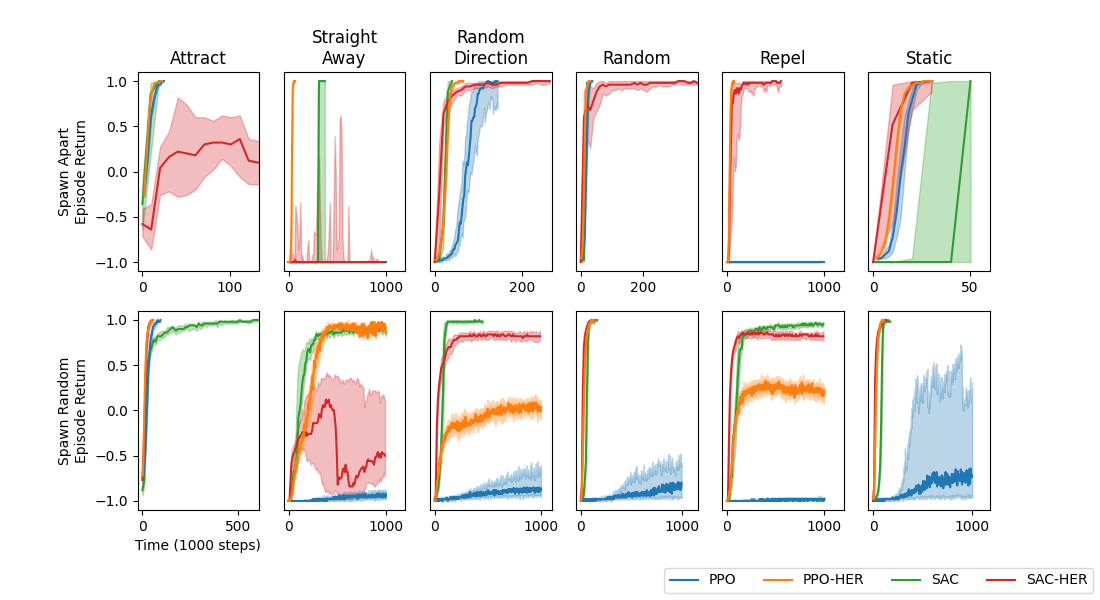}
    \caption{In general, PPO-HER achieves higher rewards than PPO, SAC, or SAC-HER while also being as sample efficient.  Bold lines and shaded regions represent the median and interquartile range, respectively.}
    \label{fig:sacHerSteps}
\end{figure*}

\begin{figure*}
    \centering
    \includegraphics[width=\textwidth]{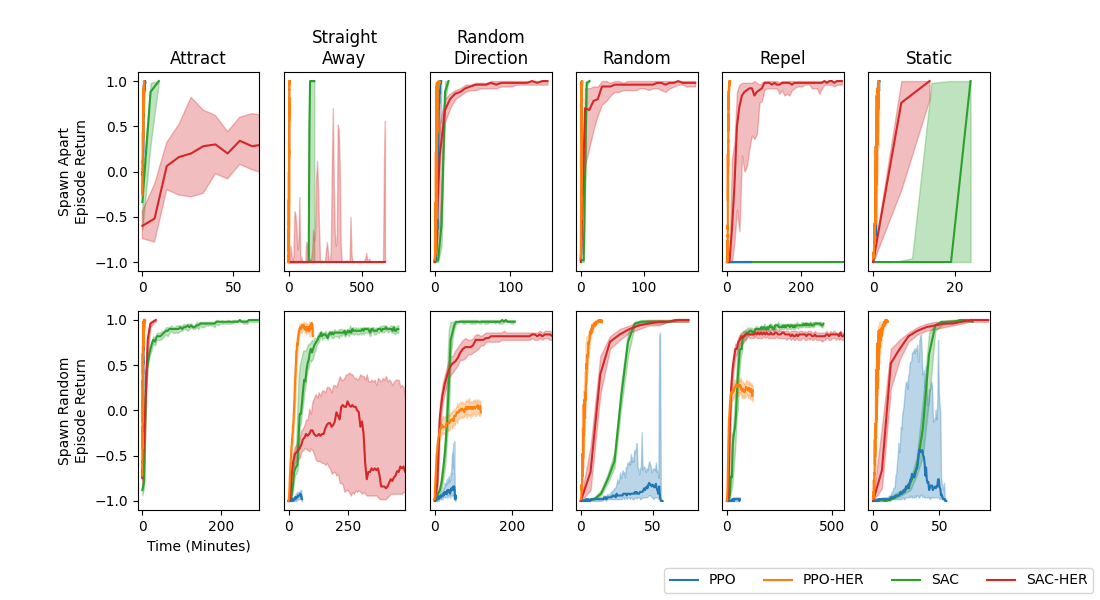}
    \caption{In general, PPO-HER achieves higher rewards than PPO, SAC, or SAC-HER while also being more clock-time efficient.  Bold lines and shaded regions represent the median and interquartile range, respectively.}
    \label{fig:sacHerTime}
\end{figure*}

Across the Predator-Prey environments (Figure \ref{fig:sacHerSteps}), PPO-HER dramatically outperformed PPO, except on the easiest environments (Attract).  For certain environments (Repel and Straight Away, particularly for the Spawn Apart condition),  PPO-HER was able to solve environments that PPO, when used without HER, was unable to solve.  Notably, HER not only accelerated PPO, but it also resulted in higher median rewards.  And, when examining the interquartile ranges, it is clear that HER stabilized RL.

When comparing PPO-HER to SAC and SAC-HER, PPO-HER achieved higher median episode returns in 10 of the 12 conditions.  Additionally, PPO-HER was the most sample efficient in 6 of the 12 conditions.  In the remaining 6 conditions, PPO-HER converged just 27\% slower (median) than the fastest algorithm (measured using number of steps).  At its worst, PPO-HER required 382\% of the time steps of the leading algorithm to converge, and at its best, PPO-HER required just 7\% of the time steps as the next fastest algorithm.  Taken together, PPO-HER converges as fast, and in many cases, faster, than comparable off-policy algorithms, as measured in time steps.  This is important behavior for environments that are computationally intensive.

When comparing PPO-HER to SAC and SAC-HER using clock time (Figure \ref{fig:sacHerTime}), instead of time steps (Figure \ref{fig:sacHerSteps}),  PPO-HER converged faster than all other algorithms in 11 of the 12 conditions (the exception being Random Spawner, Random Direction).  Thus, in environments, where simulation time is negligible compared to the time it takes to train the neural network, PPO-HER is expected to outperform PPO, SAC, and SAC-HER.

\subsection{PPO-HER \& Hyperparameter Sensitivity}

\begin{figure*}
    \centering
    \includegraphics[width=\textwidth]{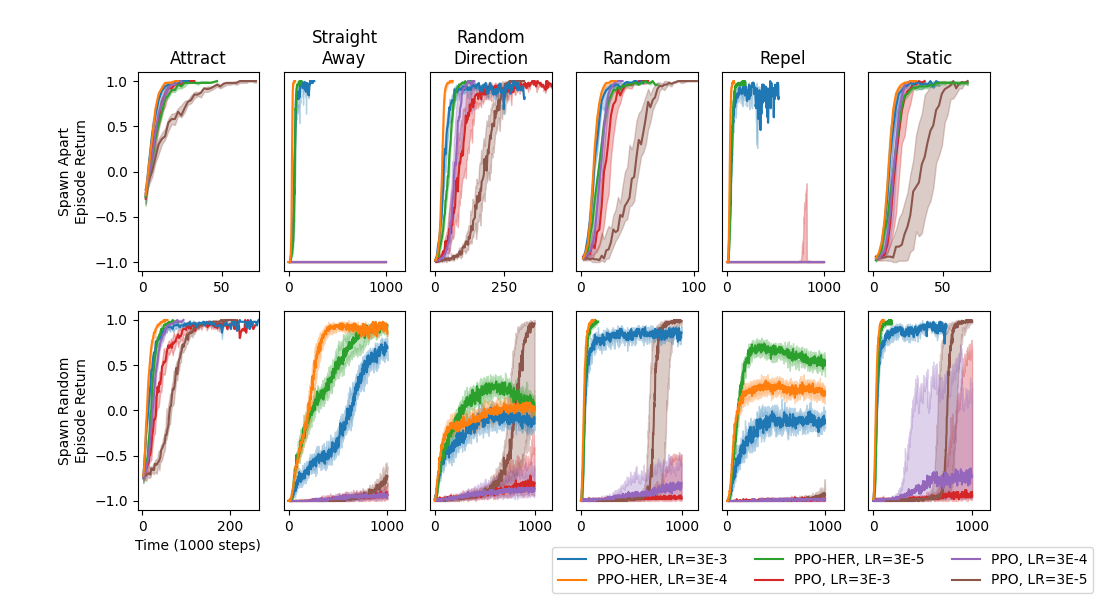}
    \caption{PPO-HER is less sensitive to hyperparameters, including the learning rate.  Bold lines and shaded regions represent the median and interquartile range, respectively.}
    \label{fig:lr}
\end{figure*}

In our initial hyperparameter tuning (limited to the Static condition), we noticed that PPO-HER seemed quite insensitive to important parameters, such as the learning rate.  To test if PPO-HER is insensitive to hyperparameters, we varied the learning rate over 3 orders of magnitude.  As shown in Figure \ref{fig:lr}, PPO-HER was relatively insensitive to changes in the learning rate; at its worst (Random Spawn, Repel), the maximum median reward changed by only 76\% between the highest-performing and lowest-performing learning rates.  PPO, at its worst (Random Spawn, Random) varied by as much as 3200\%.

\subsection{HER Scaling Benefits}
PPO-HER scales better than PPO to problems that are more difficult.

In Supplementary Figure \ref{fig:agentSize}, the interception distance was decreased, which decreased the probability that an RL agent would intercept the prey during random movements.  As the predator-prey interception distance was decreased from $I=1$ to $I=0.1$, the mean median success rate across all conditions decreased from $4.8 \times 10^{-1}$ to $8.3 \times 10^{-4}$ for PPO (99.8\% decrease in performance) and from 0.94 to 0.15 for PPO-HER (83.5\% decrease in performance).

In Supplementary Figure \ref{fig:gridSize}, the size of the environment was increased, which increased the exploration space, increased the initial distance between the predator and prey, and increased the probability that the prey could evade the predator.  As the predator-prey environment size was increased from $S=10$ to $S=100$, the mean median success rate across all conditions decreased from 0.55 to 0.099 for PPO (82.1\% decrease in performance) and from 0.88 to 0.27 for PPO-HER (69.6\% decrease in performance).

In Supplementary Figures \ref{fig:envDimHer} and \ref{fig:envDimNoHer}, the number of physical dimensions of the environment was increased, which increased the exploration space exponentially.  As the predator-prey environment dimensionality was increased from $D=2$ to $D=6$, the mean median success rate across all conditions decreased from 0.83 to 0.083 for PPO (90.0\% decrease in performance) and from 0.97 to 0.62 for PPO-HER (36.0\% decrease in performance).

\subsection{Fetch}

\begin{figure*}
    \centering
    \includegraphics[width=\textwidth]{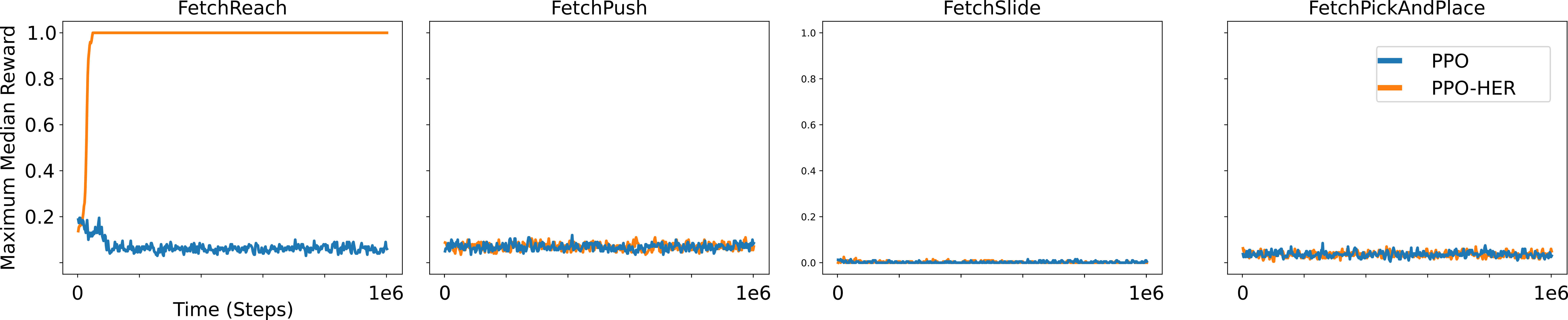}
    \caption{Results for Fetch environments.  Bold lines and shaded regions represent the median and interquartile range, respectively.}
    \label{fig:gym}
\end{figure*}

As shown in Figure \ref{fig:gym}, PPO-HER was able to learn only the FetchReach task, where it achieved a maximum success rate of 1 within 50,000 timesteps.  PPO was not able to learn any of the Fetch tasks.

%% file: source/_g_discussion.tex
\section{Discussion}

\subsection{Predator-Prey Environments}
When tested in the Predator-Prey environments, PPO-HER generally outperformed PPO, SAC, and SAC-HER, suggesting that a naive implementation of PPO-HER can be effective at solving Predator-Prey tasks.  Additionally, PPO-HER was less sensitive to the choice of the learning rate, suggesting that PPO-HER can be less sensitive to the choice of hyperparameters.  When problem complexity was increased by 1) decreasing the agent size, 2) increasing the environment size, or 3) increasing the environment dimensionality, PPO-HER greatly outperformed PPO.

In several conditions, PPO-HER did not converge to the optimal reward (e.g., Random Direction, Random Spawn).  Future work should consider ways to improve PPO-HER to allow it to achieve higher rewards for these more challenging environments.  It may be possible to improve results by modifying HER to consider the fact that goals in the Predator-Prey environments are dynamic (moving), as done in \cite{fang2018dher}.

\subsection{Fetch Environments}
When tested on the Fetch environments, PPO-HER outperformed PPO but was unable to learn most environments.  There are several possible reasons for this failure.  Firstly, many of the Fetch environments require the robotic arm to move a block.  Random motions of the arm will not cause the block to move very frequently, meaning that the achieved goal (the position of the block) does not vary much.  Under these circumstances, it is likely that off-policy algorithms, such as SAC, can take advantage of the sparse successes, whereas on-policy algorithms, like PPO may tend to overfit to setting the desired goal to be the achieved goal.  When the block is not moved, PPO-HER will tend to move the desired goal to the achieved goal and then reward the agent for not disturbing the block (i.e., not moving the block away from the re-sampled desired goal).  When combined with small rollout buffers (as opposed to the larger replay buffers used in off-policy algorithms), this may make PPO more likely to learn to move the arm away from the block instead of towards it.  This may explain why trajectory filtering, as used in \cite{dai2020episodic}, may work when combined with PPO.

\subsection{HER and On-Policy Algorithms}
For many years, there has been interest in applying HER to on-policy algorithms \cite{plappert2018multi}, although conventional wisdom says that this process should not be straightforward since ``on-policy'' implies that the data being used to update the policy was generated by the policy, and HER violates this assumption by using data that was not generated by the policy.  However, there is mounting evidence that HER can be used in on-policy algorithms. \cite{rauber2017hindsight} demonstrated that HER can be applied to policy gradients, at least for simple discrete environments.  \cite{zhang2019hindsight} showed that HER can be applied to TRPO.  \cite{iql} showed that HER can be applied to PPO via IQL.  \cite{dai2020episodic} showed that HER can help PPO by applying it indirectly to generate expert trajectories for self-imitation learning.  This study demonstrated that HER could be directly applied to PPO without any adaptation, at least for simple Predator-Prey environments.

Given this increasing body of evidence, it is interesting to consider why PPO-HER out-performs PPO, despite violating the on-policy assumption.  We believe that PPO may be compatible with HER because it uses Gaussian distributions to produce stochastic actions.  Because Gaussian distributions have infinite domains, there are many actions that are improbable, given a policy, but there are no actions that are impossible.  Given this interpretation, we can think of HER as an efficient way to sample many, many times and then filter the sampled data to select particularly informative samples (successes) for training.  We consider this conceptual interpretation to be one of the most important contributions of this work, since it justifies inquiry into how other off-policy methods for increasing sample efficiency can be applied to PPO.

In support of this conceptual interpretation, it is important to note that we observed that the stability of PPO-HER diminished as the policy improved.  We were able to stabilize PPO-HER for high success rates by tuning the target kl parameter and using early stopping.  However, for more difficult environments (e.g., as shown in Supplementary Figure \ref{fig:gridSize}), PPO-HER can be seen to result in a type of catastrophic forgetting once the agent is mostly successful.  Generally, improvements in PPO performance are associated with decreases in policy entropy.  In the limit, as the policy entropy decreases to 0, the policy becomes deterministic, and the proposed Gaussian justification for using PPO with HER becomes invalid.  When calculating PPO gradients, one of the key terms involves calculating the ratio of probabilities that an action will be chosen under the current policy and a previous policy.  As the standard deviation of the Gaussian distribution approaches zero, the ratio of probabilities will frequently be either 0 or undefined (divided by zero), thereby destabilizing PPO training.

However, an alternative interpretation of the evidence could be that violations of the on-policy assumption are not detrimental to on-policy deep RL algorithms.  It may be that neural networks and their associated training methods are robust to departures from the assumptions of on-policy algorithms.  Or, as is seen in other domains, solutions that are derived with certain assumptions may be robust to deviations from those assumptions \cite{boneau1960effects}.  Future studies should explore these possibilities.

%% file: source/_h_conclusions.tex
\section{Conclusions}
PPO-HER outperforms PPO, SAC, and SAC-HER in many different Predator-Prey environments, and the performance of PPO-HER improves, compared to PPO, as the environment difficulty increases.  However, PPO-HER does not work for most Fetch environments, and future work should consider how to generalize PPO-HER to environments when most trajectories do not result in movement of the achieved goal.

%% file: source/_a_acknowledgements.tex
The authors would like to thank Megan Emmons and Srideep Musuvathy for helpful discussion.
This work was supported by the Laboratory Directed Research and Development program at Sandia National Laboratories, a multimission laboratory managed and operated by National Technology and Engineering Solutions of Sandia LLC, a wholly owned subsidiary of Honeywell International Inc. for the U.S. Department of Energy's National Nuclear Security Administration under contract DE-NA0003525. The authors own all right, title and interest in and to the article and is solely responsible for its contents. The United States Government retains and the publisher, by accepting the article for publication, acknowledges that the United States Government retains a non-exclusive, paid-up, irrevocable, world-wide license to publish or reproduce the published form of this article or allow others to do so, for United States Government purposes. The DOE will provide public access to these results of federally sponsored research in accordance with the DOE Public Access Plan https://www.energy.gov/downloads/doe-public-access-plan.  This paper describes objective technical results and analysis. Any subjective views or opinions that might be expressed in the paper do not necessarily represent the views of the U.S. Department of Energy or the United States Government.

%% file: source/_i_supplement.tex
\section{Supplementary Information}

\begin{figure*}
    \centering
    \includegraphics[width=\textwidth]{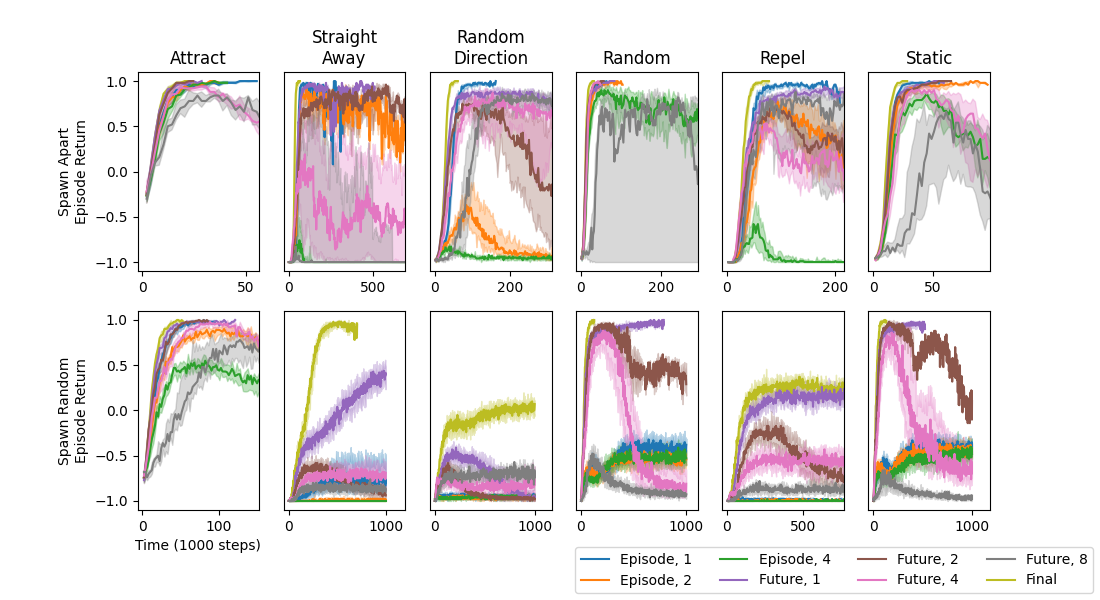}
    \caption{Hyperparameter optimization for PPO-HER. The ``final'' method for selecting new goals ($g'$) performed the best.  Note that the ``final'' method does not rely on a $k$ parameter. Bold lines and shaded regions represent the median and interquartile range, respectively.}
    \label{fig:herType}
\end{figure*}

\begin{figure*}
    \centering
    \includegraphics[width=\textwidth]{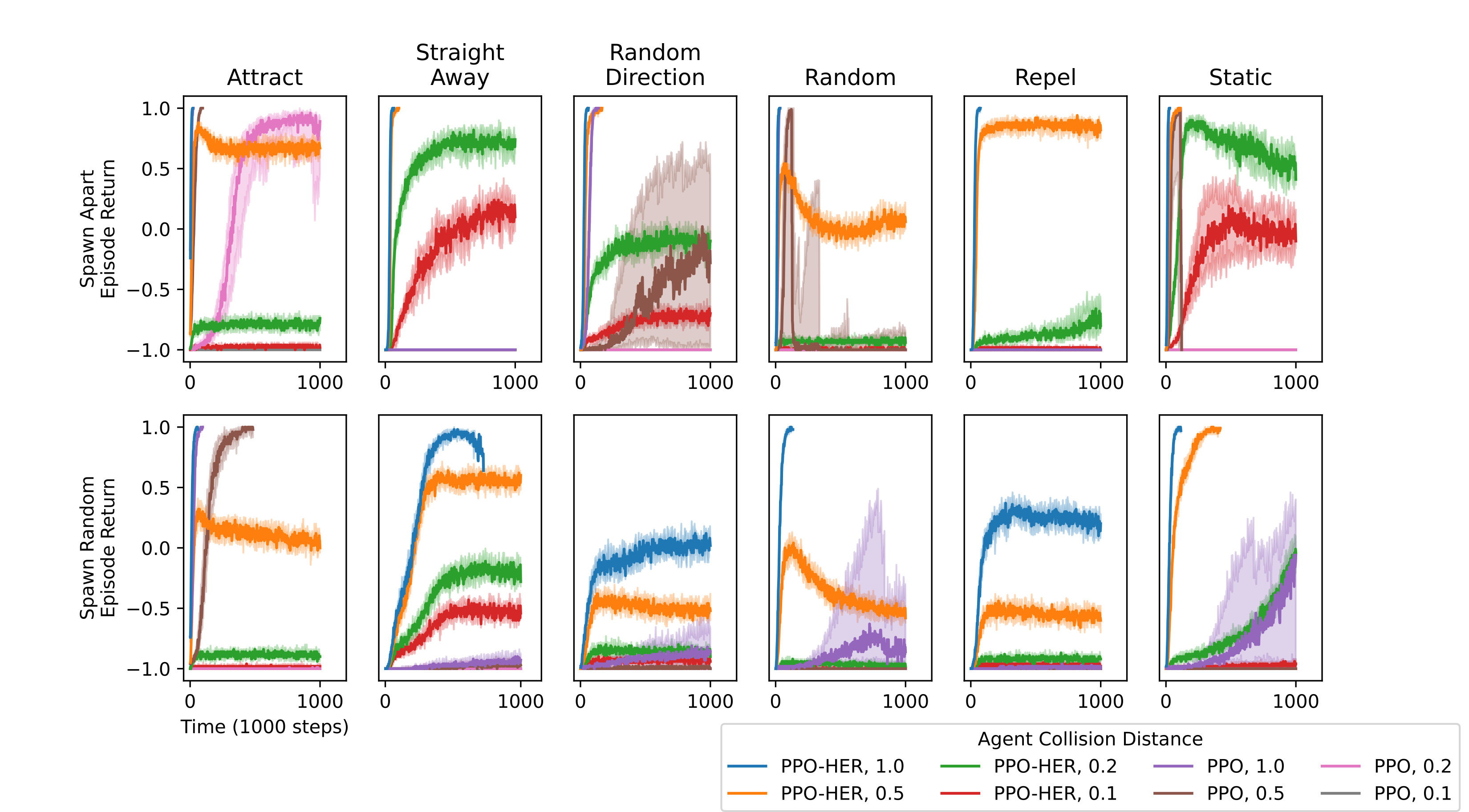}
    \caption{The interception distance was decreased from 1 to 0.1 arbitrary units.  PPO-HER generalized to smaller interception distances (harder environments) better. Bold lines and shaded regions represent the median and interquartile range, respectively.}
    \label{fig:agentSize}
\end{figure*}

\begin{figure*}
    \centering
    \includegraphics[width=\textwidth]{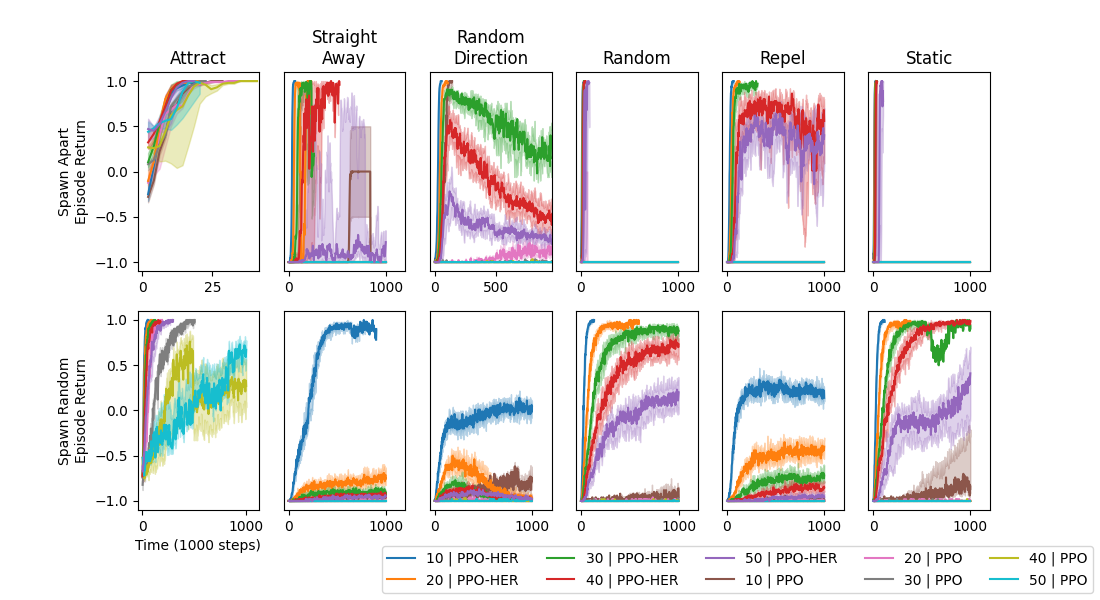}
    \caption{Environment size was increased from 10 to 50 arbitrary units.  PPO-HER generalizes to larger environment sizes (harder environments) better. Bold lines and shaded regions represent the median and interquartile range, respectively.}
    \label{fig:gridSize}
\end{figure*}

\begin{figure*}
    \centering
    \includegraphics[width=\textwidth]{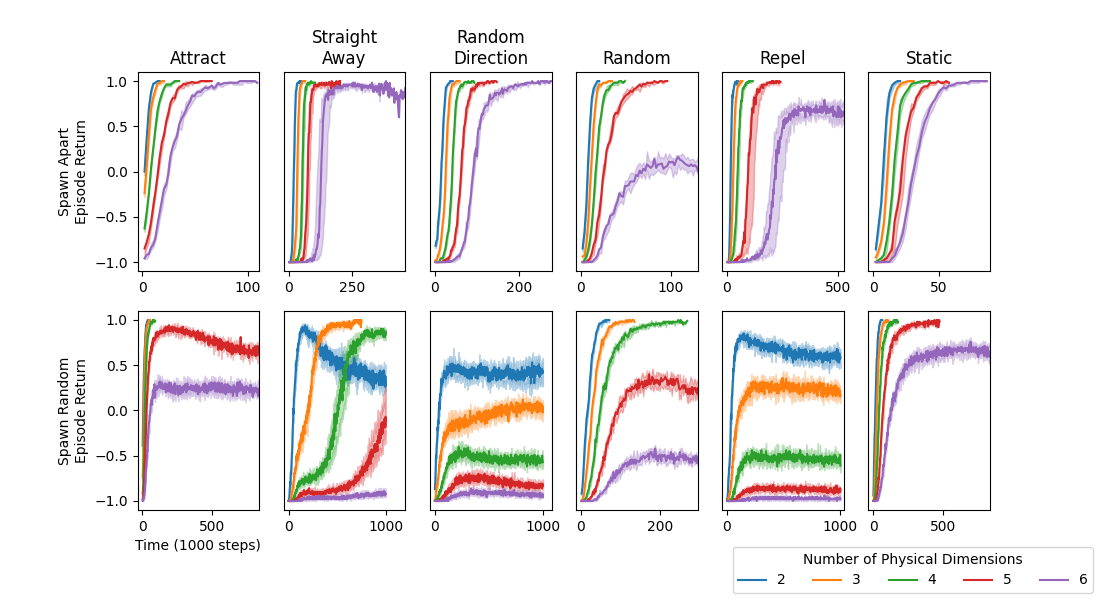}
    \caption{PPO-HER performed well when the dimensionallity of the environment was increased from 2 to 6.  See Figure \ref{fig:envDimNoHer} for results when using PPO, only.  Bold lines and shaded regions represent the median and interquartile range, respectively.}
    \label{fig:envDimHer}
\end{figure*}

\begin{figure*}
    \centering
    \includegraphics[width=\textwidth]{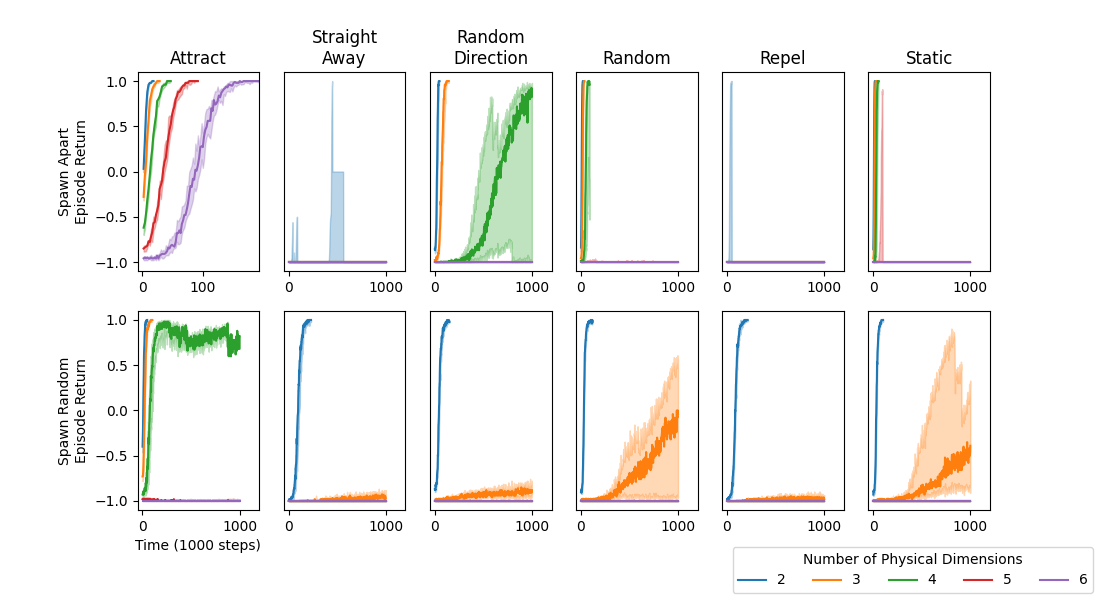}
    \caption{PPO performance was severly impaired when the dimensionallity of the environment was increased from 2 to 6.  See Figure \ref{fig:envDimHer} for results when using PPO-HER.  Bold lines and shaded regions represent the median and interquartile range, respectively.}
    \label{fig:envDimNoHer}
\end{figure*}